\documentclass[11pt,a4paper]{article}
\usepackage[hyperref]{emnlp2020}
\usepackage{times}
\usepackage{latexsym}

\usepackage{multirow}
\usepackage{multicol}
\usepackage{amsmath}
\usepackage{xcolor}
\usepackage{graphicx}
\usepackage{array}
\usepackage{booktabs}
\usepackage{covington}
\usepackage{soul}
\usepackage{subcaption}
\usepackage{bbm}
\usepackage[linesnumbered,ruled,vlined]{algorithm2e}

\SetCommentSty{mycommfont}

\definecolor{MyColor}{RGB}{50, 100, 250}
\definecolor{Orange}{RGB}{244, 101, 66}

\usepackage{microtype}

\aclfinalcopy % Uncomment this line for the final submission
\def\aclpaperid{2100} %  Enter the acl Paper ID here

\title{Incorporating a Local Translation Mechanism \\into Non-autoregressive Translation}

\author{
	Xiang Kong\bf{\thanks{\quad Zhisong and Xiang contributed equally for this paper}}, Zhisong Zhang\footnotemark[1], Eduard Hovy\\
	Language Technologies Institute, Carnegie Mellon University\\
	\texttt{\{xiangk,zhisongz,hovy\}@cs.cmu.edu}
}

\date{}

\begin{document}
\maketitle
\begin{abstract}

In this work, we introduce a novel local autoregressive translation (LAT) mechanism into non-autoregressive translation (NAT) models so as to capture local dependencies among target outputs. Specifically, for each target decoding position, instead of only one token, we predict a short sequence of tokens in an autoregressive way. We further design an efficient merging algorithm to align and merge the output pieces into one final output sequence. We integrate LAT into the conditional masked language model (CMLM; ~\citealp{ghazvininejad-etal-2019-mask}) and similarly adopt iterative decoding. Empirical results on five translation tasks show that compared with CMLM, our method achieves  comparable or better performance with fewer decoding iterations, bringing a 2.5x speedup. Further analysis indicates that our method reduces repeated translations and performs better at longer sentences. The code for our
model is available at \url{https://github.com/shawnkx/NAT-with-Local-AT}.
\end{abstract}

\section{Introduction}
\label{sec:intro2}

Traditional neural machine translation (NMT) models~\cite{sutskever2014sequence,cho2014learning,bahdanau2014neural,gehring2017convolutional,vaswani2017attention} commonly make predictions in an incremental token-by-token way, which is called autoregressive translation (AT).
Although this strategy can capture the full translation history, it has relatively high decoding latency.
To make the decoding more efficient, non-autoregressive translation (NAT)~\cite{gu2017non} is introduced to generate multiple tokens at once instead of one-by-one. 
However, with the conditional independence property~\cite{gu2017non}, NAT models do not directly consider the dependencies among output tokens, which may cause errors of repeated translation and incomplete translation~\cite{wang2019non}. 
There have been various methods in previous work~\cite{stern2019insertion,NIPS2019_9297,ma-etal-2018-bag,wei2019imitation,ma2019flowseq,tu2020engine} to mitigate this problem, including iterative decoding~\cite{lee2018deterministic,ghazvininejad-etal-2019-mask}. 
% considering local dependencies \cite{li2020lava}, etc.

\begin{figure}
    \centering
    \includegraphics[width=0.4\textwidth]{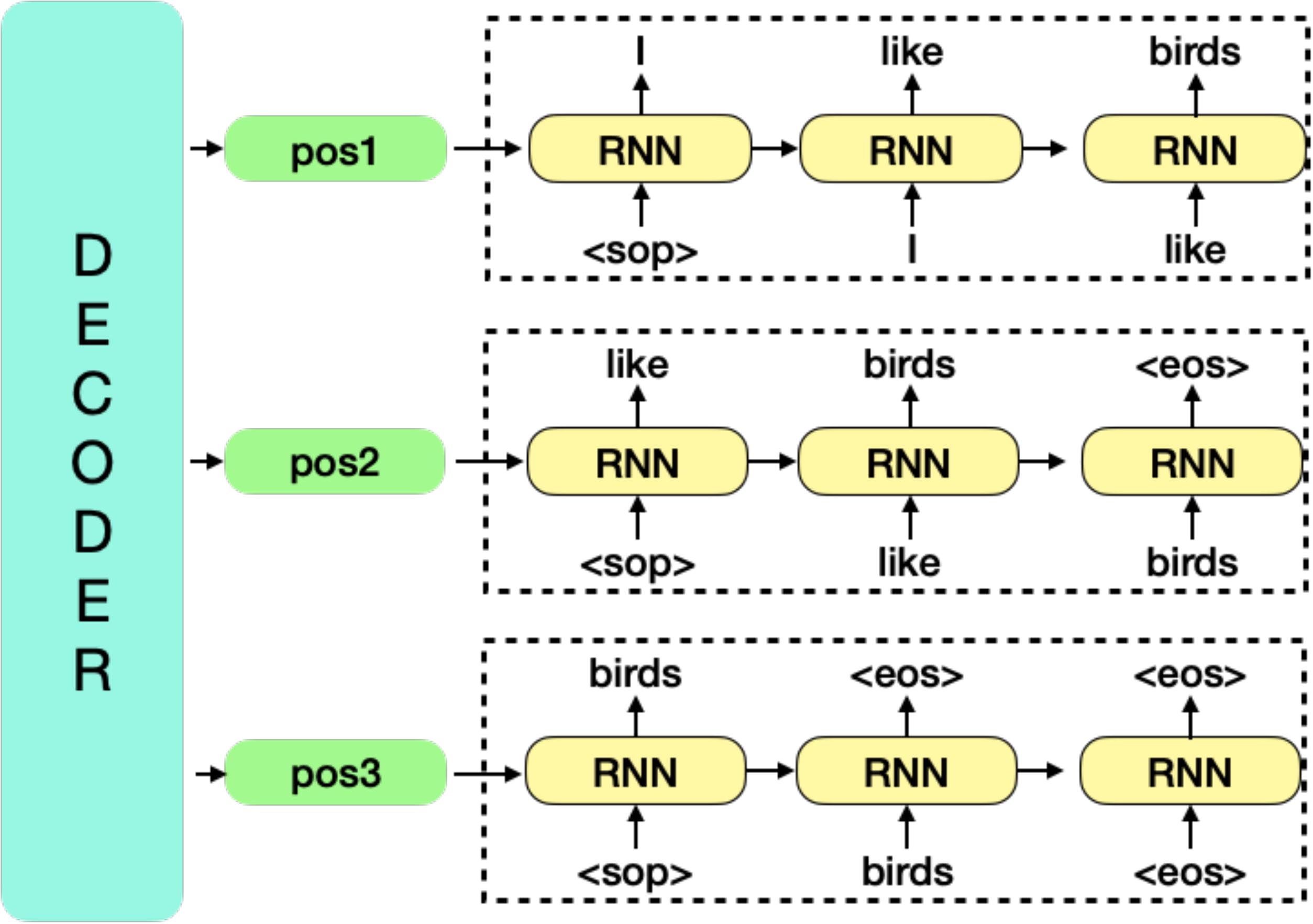}
    \caption{An example of the LAT mechanism. For each decoding position, a short sequence of tokens is generated in an autoregressive way. $\left< \textup{sop}\right>$ is the special start-of-piece symbol. `pos*' denotes the hidden state from the decoder at that position.}
    \label{fig:local_trans_schema}
    %  \vspace{-2.5ex}
\end{figure}

In this work, we introduce a novel mechanism, i.e., local autoregressive translation (LAT), to take local target dependencies into consideration. For a decoding position, instead of generating one token, we predict a short sequence of tokens (which we call a translation piece) for the current and next few positions in an autoregressive way. A simple example is shown in Figure~\ref{fig:local_trans_schema}. 

With this mechanism, there can be overlapping tokens between nearby translation pieces. We take advantage of these redundancies, and apply a simple algorithm to align and merge all these pieces to obtain the full translation output. Specifically, our algorithm builds the output by incrementally aligning and merging adjacent pieces, based on the hypothesis that each local piece is fluent and there are overlapping tokens between adjacent pieces as aligning points. 
Moreover, the final output sequence is dynamically decided through the merging algorithm, which makes the decoding process more flexible.

We integrate our mechanism into the conditional masked language model (CMLM)~\cite{ghazvininejad-etal-2019-mask} and similarly adopt iterative decoding, where tokens with low confidence scores are masked for prediction in more iterations. With evaluations on five translation tasks, i.e., WMT'14 EN$\leftrightarrow$DE, WMT'16 EN$\leftrightarrow$RO and IWSLT'14 DE$\rightarrow$EN, we show that our method could achieve similar or better performance compared with CMLM and AT models while gaining nearly 2.5 and 7 times speedups, respectively. 
Furthermore, our method is shown to effectively reduce repeated translations and perform better at longer sentences.

% \section{Conditional Masked Language Model with Local Autoregressive Translator}
\section{CMLM with LAT}
\label{sec:method}

\subsection{Model}

We integrate our LAT mechanism into CMLM, which predicts the full target sequence based on the source and partial target sequence. We adopt a lightweight LSTM-based sequential decoder as the local translator upon the CMLM decoder outputs.
For a target position $i$, the CMLM decoder produces a hidden vector $pos_i$, based on which the local translator predicts a short sequence of tokens in an autoregressive way, i.e., $t_i^1,t_i^2,...,t_i^K$. 
Here $K$ is the number of location translation steps, which is set to 3 in our experiments to avoid affecting the speed much.

\subsection{Decoding}
During inference, a special token, $\left<\textup{sop}\right>$ (start of piece) is fed into the local translator to generate a short sequence based on the $pos_{i}$.
After generating the local pieces for all target positions in parallel, we adopt a simple algorithm to merge them into a full output sequence. This merging algorithm is described in detail in Section \ref{sec:merge}.
We also perform iterative decoding following the same Mask-Predict strategy \cite{ghazvininejad-etal-2019-mask,devlin2019bert}.
In each iteration, we take the output sequence from the last iteration and mask a subset of tokens with low confidence scores by a special $\left<\textup{mask}\right>$ symbol.
Then the masked sequence is fed together with the source sequence to the decoder for the next decoding iteration.

Following~\newcite{ghazvininejad-etal-2019-mask}, a special token $\textup{LENGTH}$ is added to the encoder, which is utilized to predict the initial target sequence length. Nevertheless, our algorithm can dynamically adjust the final output sequence and we find that our method is not sensitive to the choice of target length as long as it falls in a reasonable range.

\subsection{Training}
The training procedure is similar to that of \newcite{ghazvininejad-etal-2019-mask}. Given a pair of source and target sequences $S$ and $T$, we first sample a masking size from a uniform distribution from [1, $N$], where $N$ is the target length. Then this size of tokens are randomly picked from the target sequence and replaced with the $\left<\textup{mask}\right>$ symbol. We refer to the set of masked tokens as $T_{mask}$.
Then for each target position, we adopt a teacher-forcing styled training scheme to collect the cross-entropy losses for predicting the corresponding ground-truth local sequences, the size of which is $K=3$.

Assume that we are at position $i$, we simply setup the ground-truth local sequence $t_i^1,t_i^2,...,t_i^K$ as $T_{i},T_{i+1},...,T_{i+K-1}$, where $T_{i}$ denotes the $i$-th token in the full target ground-truth sequence.
We include all tokens in our final loss, whether they are in $T_{mask}$ or not, but adopt different weights for the masked tokens that do not appear in the inputs.
Therefore, our token prediction loss function is:
{\setlength\abovedisplayskip{4pt}
\setlength\belowdisplayskip{4pt} 
\begin{align*}
\mathcal{L} =& -\sum_{i=1}^{N}\sum_{j=1}^{K}\mathbbm{1}\left\{t_{i}^{j}\in T_{mask}\right\}\log(p(t_{i}^{j}))\\
&- \sum_{i=1}^{N}\sum_{j=1}^{K}\mathbbm{1}\left\{t_{i}^{j}\notin T_{mask}\right\}\alpha\log(p(t_{i}^{j}))
\end{align*}
}
Here, we adopt a weight $\alpha$ for the tokens that are not masked in the target input, which is set as 0.1 so that the model could be trained more on the unseen tokens.
Furthermore, we randomly delete certain positions (the number of deletion is randomly sampled from [1, 0.15*$N$]) from the target inputs to encourage the model to learn insertion-styled operations. The final loss is the addition of the token prediction and the target length prediction loss. 
% The Section for the merging heuristic

\section{Merging Algorithm}
\label{sec:merge}

In decoding, the model generates local translation pieces for all decoding positions. We adopt a simple algorithm that incrementally builds the output through a piece-by-piece merging process. Our hypothesis is that if the local autoregressive translator is well-trained, then 1) the token sequence inside each piece is fluent and well-translated, 2) there are overlaps between nearby pieces, acting as aligning points for merging.

\begin{figure}[t]
	\centering
	\includegraphics[width=0.4\textwidth]{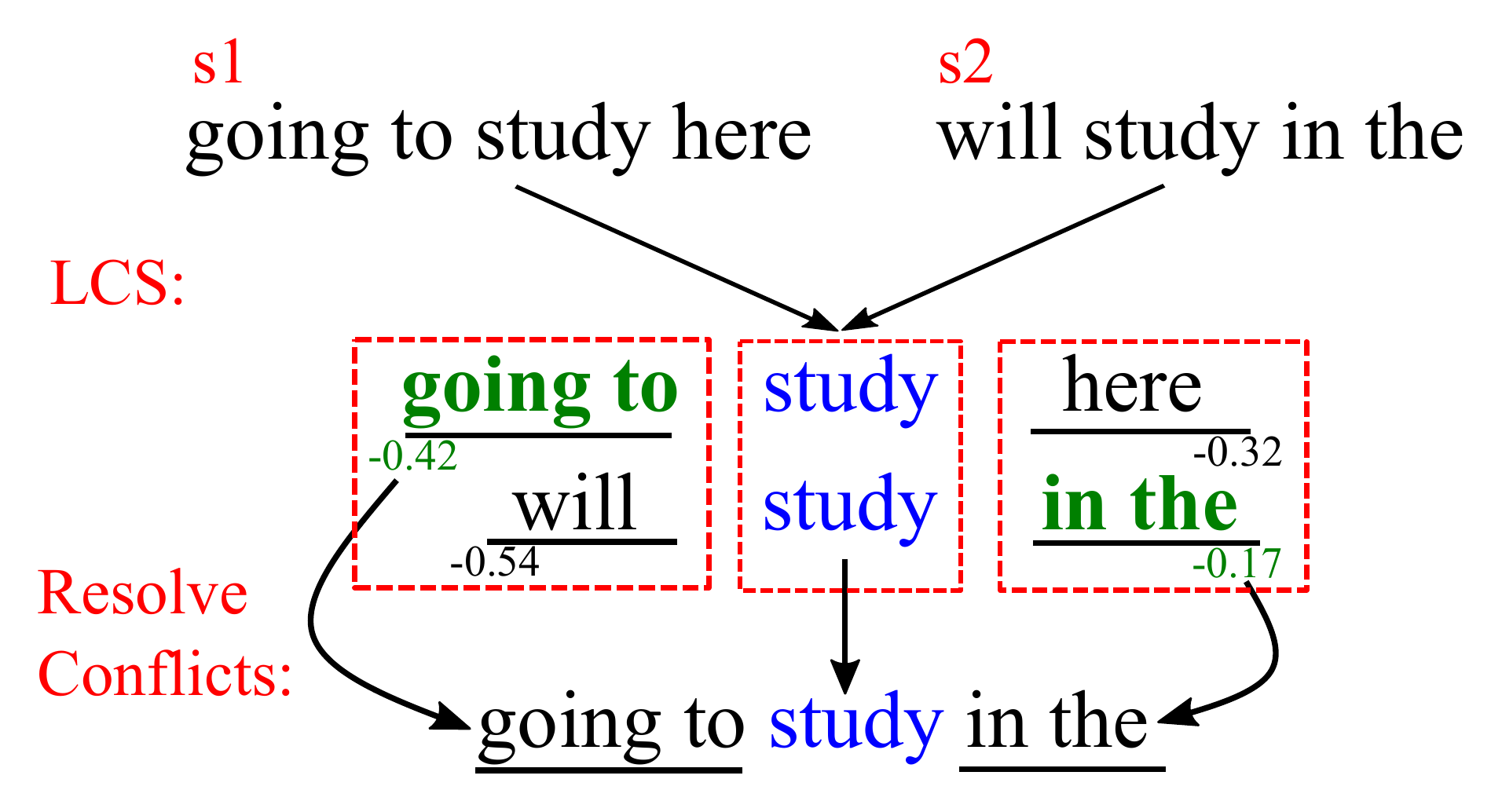}
	\caption{An example of merging two pieces of tokens.}
	\label{fig:merge0}
\end{figure}

We first illustrate the core operation of merging two consecutive pieces of tokens. Algorithm \ref{alg:merge} describes the procedure and Figure \ref{fig:merge0} provides an example. Given two token pieces $s1$ and $s2$, we first use the Longest Common Subsequence (LCS) algorithm to find matched tokens (Line 1). If there is nothing that can be matched, then we simply do concatenation (Line 3), otherwise we solve the conflicts of the alternative spans by comparing their confidence scores (Line 9-14). 
Finally we can arrive at the merged output after resolving all conflicted spans.

In the above procedure, we need to specify the score of a span. Through preliminary experiments, we find a simple but effective scheme. From the translation model, each token gets a model score of its log probability. 
For the score of a span, we average the scores of all the tokens inside. If the span is empty, we utilize a pre-defined value, which is empirically set to $\log 0.25$. For aligned tokens, we choose the highest scores among them for later merging process (Line 16).
% In the algorithm, multiple tokens can be aligned into one, among them, we adopt the highest score for later merging process.

With this core merging operation, we apply a left-to-right scan to merge all the pieces in a piece-by-piece fashion. For each merging operation, we only take the last $K$ tokens of $s1$ and the first $K$ tokens of $s2$, while other tokens are directly copied. This ensures that the merging will only be local, to mitigate the risk of wrongly aligned tokens. Here, $K$ is again the local translation step size.

Our merging algorithm can be directly applied at the end of each iteration in the iterative decoding. However, since the output length of the merging algorithm is not always the same as the number of input pieces, we further adopt a length adjustment procedure for intermediate iterations. Briefly speaking, we adjust the output length to the predicted length by adding or deleting certain amounts of special $\left< \textup{mask} \right>$ symbols. Please refer to the Appendix for more details.

Although our merging algorithm is actually autoregressive, it does not include any neural network computations and thus can run efficiently. 
In addition to efficiency, our method also makes the decoding more flexible, since the final output is dynamically created through the merging algorithm.

\begin{algorithm}[t]
    \small
	\KwIn{Two pieces of tokens: $s1$, $s2$.}
	\KwOut{A merged sequence $s'$.}
	\tcp{Call Longest Common Subsequence}
	MatchedPairs = LCS($s1$, $s2$)\;
	\If{MatchedPairs.size() == 0}{
		return $s1$+$s2$ \tcp*{Simple concat}
	}
	\Else{
		$s'$ = [] \tcp*{Initialize}
		$p1$, $p2$ = -1, -1 \tcp*{Previous idxes}
		\tcp{Add sentinel indexes.}
		MatchedPairs += [($\infty$, $\infty$)]\;
		\ForEach{$i1$, $i2$ \textbf{in} MatchedPairs}{
			$span1$ = $s1$[$p1$+1:$i1$]\;
			$span2$ = $s2$[$p2$+1:$i2$]\;
			\tcp{Solve conflicts by scores.}
			\If{score($span1$) $\geq$ score($span2$)}{
				$s'$ += $span1$\;
			}
			\Else{
				$s'$ += $span2$\;
			}
			\tcp{Align the matched ones.}
			\If{$i1 \neq \infty$}{
				$s'$ += [align($s1$[$i1$], $s2$[$i2$])]\;
			}
			$p1$, $p2$ = $i1$, $i2$\;
		}
		return $s'$\;
	}
	\caption{Merging two pieces.}
	\label{alg:merge}
\end{algorithm}

\section{Experiments}
\begin{table*}[th!]
\centering
\small
\begin{tabular}{clcllllcc}
\toprule
\multirow{2}{*}{\textbf{\#}} & \multirow{2}{*}{\textbf{Model}} & \multirow{2}{*}{\textbf{Iterations}} & \multicolumn{2}{c}{\textbf{WMT'14}} & \multicolumn{2}{c}{\textbf{WMT'16}} & \textbf{IWSLT'14} & \multirow{2}{*}{\textbf{latency (ms)}}\\
 & & &\textbf{EN-DE} & \textbf{DE-EN} & \textbf{EN-RO} & \textbf{RO-EN} & \textbf{DE-EN}\\
 \midrule
1 & AT &  $N$~~~~~~ & 27.46 & 31.87 & 34.39 & 33.98 & 34.18 & 486\\
\midrule
2 &CMLM
& \multirow{2}{*}{1~~~~~~} &  18.05 & 21.83 & 27.32 & 28.20 & 28.14 & 27\\
3 &\textbf{LAT} &   & \textbf{25.20} &\textbf{29.91}&\textbf{30.74}&\textbf{31.24}&\textbf{31.92} & \textbf{31}\\
\midrule
4 &CMLM
& \multirow{2}{*}{4~~~~~~} & 25.94 & 29.90 & 32.53 & 33.23 & 32.87 & 72\\
5 &\textbf{LAT} &  & \textbf{27.35} & \textbf{32.04} & \textbf{32.87} & \textbf{33.26} & \textbf{34.08} & \textbf{73}\\
\midrule
6 &CMLM
 & 10~~~~~~ & 27.03 & 30.53 & 33.08 & 33.31 & 33.40 & 166\\
\bottomrule
\end{tabular}
\caption{The comparisons (on BLEU score and decoding latency) of CMLM, LAT and AT models.
}
\label{tab:main-rst}
\end{table*}

\begin{table}[!ht]
    \centering
    \small
    \begin{tabular}{lccccc}
        \toprule
        \textbf{Model} & \textbf{Iteration} & \multicolumn{4}{c}{\textbf{ngram repeat rate (\%)}} \\
        & & \textbf{1} & \textbf{2} & \textbf{3} & \textbf{4}\\
        \midrule
        CMLM &  \multirow{2}{*}{1} & 20.85 & 3.78 & 1.06 & 0.37\\
        \textbf{LAT} & & \textbf{4.89} & \textbf{0.42} & \textbf{0.05} &  \textbf{0.00}\\
        \midrule
        CMLM &  \multirow{2}{*}{4} & 3.97 & 0.14 & 0.03 & 0.02\\
       \textbf{LAT} & & \textbf{3.32} & \textbf{0.08} & \textbf{0.00} & \textbf{0.00}\\ 
       \midrule
       CMLM & 10 & 3.56 & 0.08 & 0.02 & 0.02\\
        \midrule
        \midrule
        AT & $N$ & 3.27 & 0.05 & 0.00 & 0.00 \\
        Reference & - & 2.49 & 0.03 & 0.00 & 0.00\\
        \bottomrule
    \end{tabular}
    \caption{N-gram repeat rates of various models on WMT'14 EN-DE test set.
    % `AT' here is a base transformer translation model.
    }
    \label{tab:rr}
\end{table}

\begin{table}[!ht]
    \centering
    \small
    \begin{tabular}{cccccc}
        \toprule
        & \multicolumn{5}{c}{\textbf{\# local translation steps ($K$) }}\\
        & \textbf{2} & \textbf{3} & \textbf{4} & \textbf{5} & \textbf{6}\\
        \midrule
        BLEU &  32.9  &33.8 & 34.4 & 34.5 & 34.2\\
        latency (ms) & 69 & 72 & 74 & 77 & 79 \\
        \bottomrule
    \end{tabular}
    \caption{The performance of LAT models with respect to the number of local translation steps ($K$) on IWSLT'14 DE-EN test set.}
    \label{tab:lat-step}
\end{table}

\subsection{Experimental Setup}
% \paragraph{Translation Datasets}
We evaluate our proposed method on five translation tasks, i.e., WMT'14 EN$\leftrightarrow$DE, WMT'16 EN$\leftrightarrow$RO and IWSLT'14 DE$\rightarrow$EN. Following previous works~\cite{hinton2015distilling,kim-rush-2016-sequence,gu2017non,zhou2019understanding},  we train a vanilla base transformer ~\cite{vaswani2017attention} on each dataset and use its translations as the training data. The BLEU score~\cite{papineni2002bleu} is used to evaluate the translation quality. Latency, the average decoding time (ms) per sentence with batch size 1, is employed to measure the inference speed. All models' decoding speed is measured on
a single NVIDIA TITAN RTX GPU.

We follow most of the hyperparameters for the CMLM~\cite{ghazvininejad-etal-2019-mask} in the base configuration, i.e., 6 layers for encoder and decoder, 8 attention heads, 512 embedding dimensions and 2048 hidden dimensions. The LAT is an LSTM-based neural network of size 512. 
% The number of the location translation step is set as 3, i.e., generating three tokens at every position. 
Finally, we average 5 best checkpoints according to the validation loss as our final model. Please refer to the Appendix for more details of the settings. 

\subsection{Main results}
The main results are shown in Table~\ref{tab:main-rst}. Compared with CMLM at the same number of decoding iterations (row 2 vs. 3 and row 4 vs. 5), LAT performs much better while keeping similar speed, especially when the iteration number is 1. 
Note that since our method is not sensitive to predicted length, we only take one length candidate from our length predictor instead of 5 as in CMLM.

Furthermore, LAT with 4 iterations could achieve similar or better results than CMLM with 10 iterations (row 5 vs. 6) but have a nearly 2.5x decoding speedup.
\begin{figure}[!ht]
    \centering
    \includegraphics[width=0.4\textwidth]{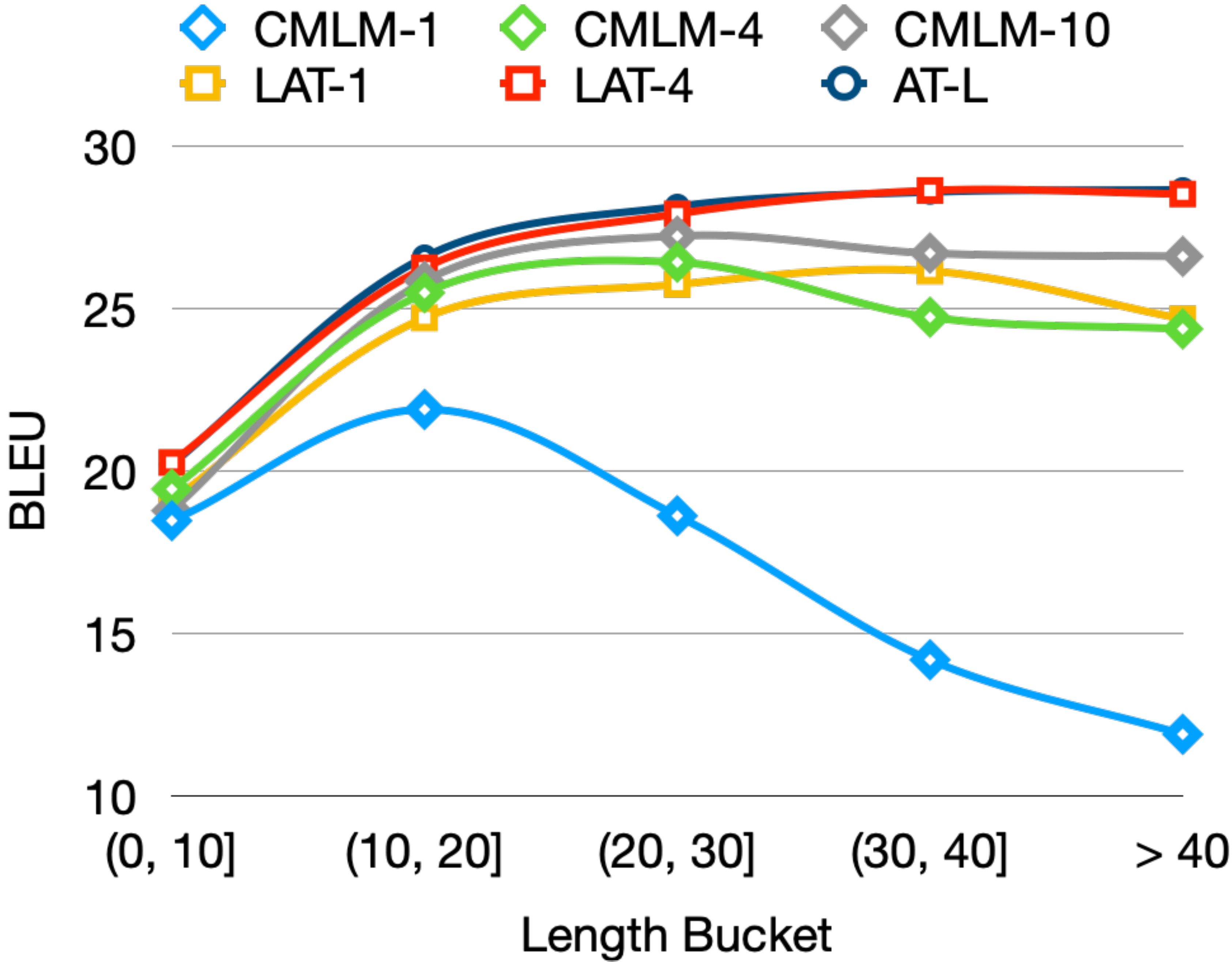}
    \caption{The BLEU scores of various systems with respect to the reference sentence lengths on WMT'14 EN-DE testset.}
    \label{fig:bleu-len-bucket}
\end{figure}
\subsection{Analysis}
\paragraph{On local translation step.}
We also explore the effects of the number of local translation steps ($K$) on the IWSLT'14 DE-EN dataset.
% Since this affects training, 
% We perform experiments on the IWSLT'14 DE-EN dataset. 
The results are shown in Table~\ref{tab:lat-step}. Generally, with more local translation steps, there can be certain improvements on BLEU but with an extra cost at inference time.
\paragraph{On repeated translation.}
We compute the $n$-gram repeat rate (nrr, what percentage of $n$-grams are repeated by certain nearby $n$-grams) of different systems on WMT'14 EN-DE test set and the result is shown in Table~\ref{tab:rr}. The nrr of CMLM with one iteration is much higher than other systems, showing that it suffers from a severe repeated translation problem. On the other hand, LAT can mitigate this problem thanks to the merging algorithm.
\paragraph{On sentence length.}
We explore how various systems perform on sentences with various lengths. The WMT'14 EN-DE test set is split into 5 length buckets by target length. Figure~\ref{fig:bleu-len-bucket} show that LAT performs better than CMLM on longer sentences, which indicates the effectiveness of our methods at capturing certain target dependencies.

\section{Related Work}
\newcite{gu2017non} begin to explore non-autoregressive translation, the aim of which is to generate sequences in parallel. In order to mitigate multimodality issue, recent work mainly tries to narrow the gap between NAT and AT. \newcite{libovicky2018end} design a NAT model using CTC loss. \newcite{lee2018deterministic} uses iteration decoding to refine translation. The conditional masked language model (CMLM)~\cite{ghazvininejad-etal-2019-mask} predicts partial target tokens based on the source text and partially masked target sentence. \newcite{ma2019flowseq} employs normalizing flows as the the latent variable to produce sequences. \newcite{sun2019fast} designs an efficient approximation for CRF for NAT.
Besides that, there are some works trying to improving the decoding speed of the autoregressive models. For example, \newcite{wang2018semi} propose a semi-autoregressive translation model, which adopts locally non-autoregressive, but autoregressive decoding. And works mentioned in~\newcite{hayashi2019findings} use techniques such as knowledge distillation, block-sparse regularization to improve the decoding speed of autoregressive models.

\section{Conclusion}

In this work, we incorporate a novel local autoregressive translation mechanism (LAT) into non-autoregressive translation, predicting multiple short sequences of tokens in parallel.
With a simple and efficient merging algorithm, we integrate LAT into the conditional masked language model (CMLM ~\citealp{ghazvininejad-etal-2019-mask}) and similarly adopt iterative decoding. We show that our method could achieve similar results to CMLM with less decoding iterations, which brings a 2.5x speedup. 
Moreover, analysis shows that LAT can reduce repeated translations and perform better at longer sentences.

\bibliographystyle{acl_natbib}
\bibliography{emnlp2020}
\clearpage
% appendix

%\onecolumn

%\appendix
%\section*{Supplementary Material}

\section*{Appendices}
\label{sec:appendix}

\newcommand{\minus}{\scalebox{0.75}[1.0]{$-$}}
\addcontentsline{toc}{section}{Appendices}
\renewcommand{\thesubsection}{\Alph{subsection}}

\subsection{Preprocessing}
We follow the standard pre-processing procedure in prior works~\cite{vaswani2017attention,lee2018deterministic}. All datasets are segmented into subwords through byte pair encoding (BPE)~\cite{sennrich2016neural}. The BPE code is learnt from the combination of source and target data for WMT datasets. For IWSLT, the bpe code is learned from the source and target data separately. Table~\ref{tab:data-pre} lists some details.
\begin{table}[!ht]
    \centering
    \begin{tabular}{lcc}
        \toprule
        Dataset &  Vocab. Size & Data size\\
        \midrule
        IWSLT & 10k & 150k\\
        WMT14 EN$\leftrightarrow$DE & 32k & 4.5M\\
        WMT16 EN$\leftrightarrow$RO & 40k & 600k\\
        \bottomrule
    \end{tabular}
    \caption{Pre-processing details of various translation benchmarks. Vocab. size denotes vocabulary size.}
    \label{tab:data-pre}
\end{table}

\begin{table*}[th!]
\centering
\small
\begin{tabular}{lcllllc}
\toprule
\multirow{2}{*}{\textbf{Model}} & \multirow{2}{*}{\textbf{Iterations}} & \multicolumn{2}{c}{\textbf{WMT'14}} & \multicolumn{2}{c}{\textbf{WMT'16}} & \textbf{IWSLT'14}\\
  & &\textbf{EN-DE} & \textbf{DE-EN} & \textbf{EN-RO} & \textbf{RO-EN} & \textbf{DE-EN}\\
 \midrule
AT &  $N$~~~~~~ & 26.13 & 31.06 & 34.74 & 35.76 & 33.59\\
\midrule
CMLM  & \multirow{2}{*}{1~~~~~~} & 18.47 & 22.83 & 26.92 & 28.77 & 24.57\\
LAT &   &22.14 & 29.20 & 32.16 & 32.07 & 28.34\\
\midrule
CMLM
& \multirow{2}{*}{4~~~~~~} & 24.73& 29.18 & 33.06 & 34.31 & 29.06\\
LAT & & 26.03 & 31.66 & 33.49 & 34.77& 34.05 \\
\midrule
CMLM
 & 10~~~~~~ & 25.25& 29.83 & 33.66 & 34.65 & 33.23\\
\bottomrule
\end{tabular}
\caption{The comparisons (on BLEU score and decoding latency) of CMLM, LAT and AT models on development sets.
% The performance (BLEU) of CMLMs with mask-predict, compared to other parallel decoding machine translation methods. The standard (sequential) transformer is shown for reference. Bold numbers indicate state-of-the-art performance among parallel decoding methods.
}
\label{tab:main-rst-dev}
\end{table*}
\subsection{Optimization}
We sample weights from $\mathcal{N}(0, 0.02)$, initialize biases to zero, and set layer normalization parameters to $\beta$ = 0, $\gamma$ = 1.
For regularization, we use 0.3 dropout, 0.01 L2
weight decay, and smoothed cross-entropy loss
with $\epsilon$ = 0.1. We train batches of 128k tokens using Adam~\cite{adam2015} with $\beta = (0.9, 0.999)$ and $\epsilon=10^{\minus 6}$. The learning rate warms up to a peak of $5\times{}10^{\minus 4}$ within 10,000 steps, and then decays with the inverse square-root schedule. 
We train our models for 300k steps with batch size 128k~\cite{ghazvininejad-etal-2019-mask} for WMT datasets. For the IWSLT dataset, we train our models for 50k steps with batch size 32k.  

\subsection{Model Parameter Size}
The averaged size of parameters for all models are shown in Table~\ref{tab:param_size}. These three kinds of models have similar number of parameters. LAT models have the most number of parameters due to the LSTM-based local translator.
\begin{table}[!ht]
    \centering
    \begin{tabular}{c|ccc}
    \toprule
        Model &  Parameter size\\
        \midrule
        AT & 60M \\
        CMLM & 62M\\
        LAT & 64M \\
        \bottomrule
    \end{tabular}
    \caption{Number of Parameters of different models.}
    \label{tab:param_size}
\end{table}
\subsection{Validation Performance}
The performance of different models on translation tasks' validation sets is reported in the Table~\ref{tab:main-rst-dev}. We could find the similar trend to the performance on the test set.

\subsection{Length Adjustment for Intermediate Iterations}

Since our merging algorithm produces the output dynamically, the output length is usually not the same as the number of input pieces. In iterative decoding, we find it helpful to adjust the output sequence's length to the input length in intermediate iterations. This is achieved by adding or deleting the special $\left< mask\right>$ symbols. Notice that for the final iteration, we do not apply any adjustments and keep the merged output sequence as it is.

For the length adjustment in the intermediate iterations, our goal is to adjust the output length of the merger ($L_{out}$) to be close to the input target length ($L_{in}$). If these two lengths are already equal or their relative difference is within a certain range (which is empirically set to 5\%), we will do nothing. Otherwise, there can be two cases: 1) when $L_{in}$ is larger than $L_{out}$, we further insert $L_{in}-L_{out}$ $\left< mask\right>$ tokens into the sequence; 2) otherwise, we try to delete $L_{out}-L_{in}$ $\left< mask\right>$ tokens. Notice that the addition or deletion operations happen after the masking procedure for the next iteration.

Here, we describe the addition case in detail. Suppose we need to further insert $M$ masks into the output sequence, we decide the insertion places according to the position gaps. 
We adopt a simple position scheme for all the tokens.
For each original token $t_i^j$ (the $j$-th token in the $i$-th piece) in the input translation pieces, we set $i+j$ as its position.
For each token in the output sequence after merging, since it can originate from multiple input tokens through aligning, we take the averaged value of all its source input tokens' positions.
We calculate the position gap between each pair of nearby unmasked tokens in the output sequence and maintain a priority queue for all these gaps. Then we insert $M$ masks once at a time. For each time, we select the current maximal gap, insert a $\left< mask\right>$ to that position, and subtract that gap by 1.
The case for deletion would be similar but in the opposite direction: select the minimal gap, delete one $\left< mask\right>$ if there are any, and increase that gap by 1.
We will delete nothing if there are no masked tokens in the selected gap.

\end{document}